\documentclass[runningheads]{llncs}
\usepackage[T1]{fontenc}
\usepackage{graphicx,verbatim}
\usepackage{amsmath}
\usepackage{amssymb}
\usepackage{booktabs}
\usepackage{graphicx}
\usepackage{multirow}
\usepackage{caption}
\usepackage{subcaption}
\usepackage{todonotes}
\usepackage{marvosym}
\usepackage{float}
\begin{document}

\title{Physiology and Anatomy Aware Inverse Inference of Myocardial Infarction for Cardiac Digital Twin\thanks{
This arXiv version corresponds to the submitted version. 
The final version will be available on Springer Link.}
}

\titlerunning{Physiology and Anatomy Aware Inverse MI Inference}
%

\author{
Mengxiao Wang\inst{1,2}, 
Yilin Lyu\inst{2}, 
Julia Camps\inst{3}, 
Ching Hui Sia\inst{2}, 
Mark Yan-Yee Chan\inst{2}, 
Yanrui Jin\inst{1}, 
Shuzhi Sam Ge\inst{2}, 
Chengliang Liu\inst{1}, 
Lei Li\inst{2}\textsuperscript{\Letter}}

\authorrunning{Wang et al.}
\institute{
Shanghai Jiao Tong University, Shanghai, China \\
\and
National University of Singapore, Singapore\\
\email{lei.li@nus.edu.sg}
\and
Universitat Pompeu Fabra, Barcelona, Spain
}
  
\maketitle              
\begin{abstract}
Accurate localization of myocardial infarction is essential for risk stratification. While LGE-MRI remains the gold standard, it is resource-intensive. Integrating cine MRI with ECG enables a more detailed representation of infarct properties. Existing inverse MI inference methods overlook realistic scar morphology and cardiac repolarization, reducing sensitivity to subtle ECG variations and interpretability of infarct-induced electrophysiological changes. In this paper, we propose a novel framework for noninvasive MI localization using cardiac digital twins. To bridge the domain gap between simulation and reality, we introduce an anatomy-aware stochastic infarct synthesis strategy to synthesize realistic, irregular scars with border zones, mimicking ischemic transmural progression. We then construct a virtual cohort to simulate QRS-T waveforms, capturing both depolarization and repolarization of electrophysiological dynamics. Furthermore, we design a Physiology and Anatomy Aware Network (PAA-Net) that jointly encodes 3D myocardial geometry and multi-lead ECGs to infer infarct area with various localizations, sizes, spatial extents and transmuralities. Experimental results demonstrate that our framework significantly outperforms existing methods in inverse inference, achieving 0.7391 and 0.5503 in $Dice_{Scar}$ and $Dice_{BZ}$, respectively, and further enhances interpretability of the ECG–infarct relationship. Our code will be released upon acceptance.


\end{abstract}

\section{Introduction}
Myocardial infarction (MI) is a major global cardiovascular disease, associated with substantial morbidity and mortality~\cite{mendisWorldHealthOrganization2011}. Post-infarction structural remodeling, particularly myocardial scar formation, constitutes a critical substrate for life-threatening arrhythmias. Consequently, accurate localization of the infarcted myocardium is essential for effective risk stratification and therapeutic planning~\cite{worldhealthorganizationWorldHealthStatistics2021}, and in cardiac digital twins, it is formulated as an inverse problem~\cite{liSolvingInverseProblem2025}.

In out-of-hospital or resource-limited settings, accurate MI diagnosis and localization remain challenging due to the trade-off between accessibility and diagnostic resolution~\cite{muCardiacTransmembranePotential2021,xuVariationalBayesianElectrophysiological2014}. Although Late Gadolinium Enhancement Magnetic Resonance Imaging (LGE-MRI) is the gold standard for MI assessment, it is costly, time-consuming, and associated with potential safety risks~\cite{trankleSyntheticContrastFreeLGE2026}. Electrocardiography (ECG) is widely accessible and convenient, it offers only limited localization capability and cannot quantify detailed infarct characteristics, such as size, spatial extent, or transmurality~\cite{christianLimitationsElectrocardiogramEstimating1991}. Consequently, a joint framework that combines MRI and ECG is indispensable for comprehensive MI assessment.

Recently, cine MRI, a standard contrast-free modality, has been explored for MI assessment by analyzing cardiac motion~\cite{xuContrastAgentfreeSynthesis2020,zhangDeepLearningDiagnosis2019}. However, most studies rely on 2D single-view representations, limiting their ability to capture the complex 3D structure of infarcts.
With the rise of cardiac digital twins, precise 3D anatomical and functional modeling~\cite{niedererComputationalModelsCardiology2019} is increasingly needed, motivating the shift from planar imaging to volumetric modeling~\cite{corral-aceroDigitalTwinEnable2020}.
Studies show that 3D motion patterns derived from cine MRI improve MI detection~\cite{pengAnatomicalSignificanceAwareArchitecture2026}, and integrating computational cardiac models enables noninvasive identification of infarct location and extent~\cite{mehdiNonInvasiveDiagnosisChronic2025}.
Recent work has combined cine MRI with ECG to infer myocardial tissue properties~\cite{liEnablingCardiacDigital2024}, and computational electrophysiology helps study infarct effects on simulated ECGs. Nevertheless, these approaches mainly address ventricular depolarization, neglecting T-wave repolarization. Moreover, irregular and stochastic scar synthesis within digital twins should account for real-world characteristics~\cite{ramzanCLAIMClinicallyGuidedLGE2026}, yet this aspect remains insufficiently addressed in existing methods. Furthermore, the inverse inference model can be enhanced to capture subtle variations in ECG morphology and to improve the interpretability of how infarcted regions influence the electrical signals.

In this work, we propose a novel framework for inverse MI inference based on cardiac digital twins, enhancing both stochastic scar synthesis and the fusion network for infarct localization. To the best of our knowledge, this is the first study to simulate full QRS–T complexes and integrate them into an inverse MI inference framework. The main contributions are summarized as follows:
\begin{enumerate}
  \item We propose an anatomy-aware stochastic strategy to generate realistic irregular scars with border zones, mimicking ischemic transmural progression and covering diverse locations, sizes, spatial extents and transmuralities.
  \item We construct a virtual cohort of QRS-T waveforms, including both depolarization and repolarization phases of the ventricle under various MI scenarios.
  \item We develop the Physiology and Anatomy Aware Network (PAA-Net) that jointly models spatial point cloud information and ECG signals for MI inverse inference, and also improves interpretability.
\end{enumerate}

\section{Method}
\begin{figure}[t]
    \centering
    \includegraphics[width=\textwidth]{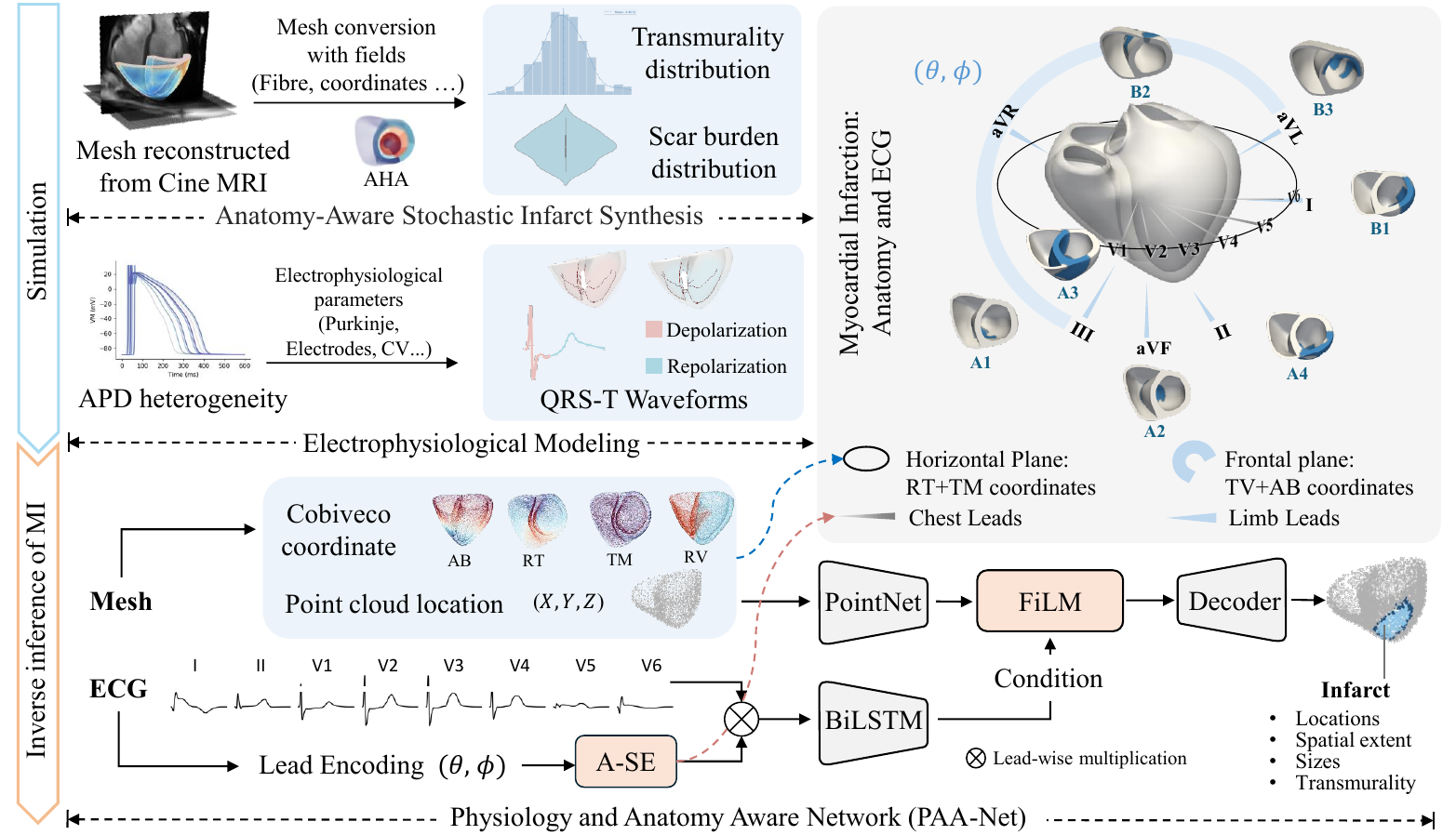}
    \caption{The pipeline of cohort simulation and inverse inference of infarcted area.}
    \label{pipeline}
\end{figure}

The overall pipeline comprises two stages as illustrated in Fig.~\ref{pipeline}. The first stage generates diverse MI scenarios through stochastic infarct synthesis, modeling variations in transmurality and scar burden while capturing ventricular depolarization and repolarization. The second stage introduces the Physiology and Anatomy Aware Network (PAA-Net) for inverse MI inference, which incorporates both the physiological characteristics of ECG lead positions and the anatomical spatial localization of infarcted regions.

\subsection{Anatomy-Aware Stochastic Infarct Synthesis}

We first construct simulation-ready biventricular meshes from multi-view cine MRI. The surface meshes are reconstructed from SAX and LAX segmentation labels following~\cite{dillonOpenSourceEndtoEndPipeline2025}, and volumetric meshes are then generated using Iso2Mesh~\cite{tranImprovingModelbasedFunctional2020}. Then, fields are assigned, including ventricular coordinates via Cobiveco~\cite{schulerCobivecoConsistentBiventricular2021}, as well as rule-based fibre orientations and electrode placements~\cite{dosteAutomatedComputationalPipeline2025}.

For scar synthesis, we considered the realistic scar distribution in Myocardial Infarction. Specifically, we  applied clinical knowledge to split the scar area into 7 types~\cite{nikusUpdatedElectrocardiographicClassification2014} based on the 17-segment  according to American Heart Association (AHA) standard. Different from~\cite{liEnablingCardiacDigital2024}, we do not consider scars as regular ellipsoid, instead, we developed a stochastic generation framework based on ventricular coordinates. The scar generation process accounts for both the spatial heterogeneity and the transmural progression characteristic of ischemic infarct. First, a spatially correlated noise field $N(x)$ was generated by applying a Gaussian filter to a random uniform distribution. Second, to mimic the transmural ischemic wavefront, the scar core was defined by a dynamic thresholding function dependent on the transmural coordinate $T_{m}$ (where $T_{m}=1$ at the endocardium):
\begin{equation}
Scar=\{x|N(x)>\tau_{base}+\lambda(1-T_{m}(x))\},
\end{equation}
where $\tau_{base}$ controls overall scar volume and $\lambda$ penalizes scar formation towards the epicardium. Finally, the border zone (BZ) was generated morphologically using a K-D tree search, reclassifying all viable nodes within a Euclidean radius $r=2\,mm$ of the scar core as BZ tissue, thereby creating a transitional halo surrounding the dense core.
The final synthetic scars contain 17 scenarios covering 7 infarct locations, 2 infarct sizes, and 2 transmurality types.

\subsection{Electrophysiological Modeling of  QRS-T Waveforms}

For electrophysiology simulation, we utilize the reaction-Eikonal model~\cite{neicEfficientComputationElectrograms2017}, which combines the Eikonal-type model with an action potential shape model to recover transmembrane voltages.
First, local activation times $t_a(x)$ are computed by solving the Eikonal equation:
\begin{equation}
\sqrt{\nabla t_a(x)^T \mathbf{V} \nabla t_a(x)} = 1, \quad x \in \Omega
\end{equation}
subject to $t_a(y_i) = t_i$ for $N_{root}$ early activation sites, where $\mathbf{V}$ represents the orthotropic conduction velocity tensor aligned with the myocardial fiber architecture.To simulate the transmembrane potential $U$, the activation wavefront $t_a(x)$ triggers a stimulus current $I_{foot}(x,t)$ over a duration $T_{foot}$. The evolution of $U$ is then governed by a diffusion-less reaction-Eikonal formulation:
\begin{equation}
\frac{\partial U}{\partial t} = -\frac{1}{C_m} \left[ I_{ion}(U, w, c) + I_{foot}(x, t) \right]
\end{equation}
where $I_{ion}$ is derived from the Mitchell-Schaeffer model to capture cellular action potential (AP) dynamics.

To incorporate repolarization gradients, we implement Action Potential Duration (APD) heterogeneity. A spatial gradient field $q(x) = g_{ab}ab(x) + g_{tm}tm(x)$ is defined to account for apicobasal ($ab$) and transmural ($tm$) variations. The local $APD(x)$ is then linearly mapped between $APD_{min}$ and $APD_{max}$ based on $q(x)$.
Finally, multi-lead ECGs are generated via the pseudo-ECG formulation, capturing the QRS complex and T-wave morphology. All signals are normalized relative to R-wave progression patterns observed in clinical cohorts~\cite{campsHarnessing12leadECG2024}. 


\subsection{Inverse inference of Myocardial Infarction}
\subsubsection{Problem Formulation.}
We assume that a given dataset $\mathcal{D}=\{(X_{i}, S_{i}, Y_{i})\}_{i=1}^{N}$ consists of $N$ subjects. The geometric input $X_{i} \in \mathbb{R}^{V \times 7}$ is constructed by concatenating the 3D Cartesian coordinates and the 4D Cobiveco ventricular coordinates for $V=4096$ nodes. The temporal input $S_{i} \in \mathbb{R}^{T \times 8}$ consists of leads I, II, and V1–V6, which provide a non-redundant representation of the cardiac electrical activity with $T=512$ timestamps. The ground truth pathology $Y_{i} \in \{0,1\}^{V \times 3}$ represents node-wise classes labels (normal, scar, and BZ).

\subsubsection{Physiology and Anatomy Aware Network (PAA-Net).} 
To effectively extract geometric features and localize infarct regions, the mesh point cloud is encoded using PointNet++~\cite{qiPointNetDeepHierarchical2017}, which is defined as $F_{pc} = \mathrm{PointNet}(X)$,
with 3D Cartesian coordinates and Cobiveco ventricular coordinates of each point.
To exploit the spatial topology of ECG electrodes, we propose the Anatomical Squeeze-and-Excitation (A-SE) module. Unlike standard SE blocks, A-SE introduces a geometric inductive bias by coupling lead-wise signal dynamics with Spherical Lead Encoding. Specifically, for each lead $l$, we compute its temporal energy $s_l = \frac{1}{T} \sum_{t} |S_{l,t}|$ and concatenate it with the lead position prior $PE_l$ derived from its physical coordinates $(\theta, \phi)$. The recalibration weight $w_l$ is then generated via a gating mechanism:
\begin{equation}
w_l = \sigma(\text{MLP}([s_l \parallel PE_l]))
\end{equation}
The final representation is obtained by $F_{ecg} = \mathrm{BiLSTM}\left(\{ w_l \cdot S_l \}_{l=1}^{L}\right)$ using the BiLSTM module~\cite{wangBioCrossCrossmodalFramework2025} for temporal feature extraction. By fusing transient signal intensity with the electrophysiological context, A-SE dynamically prioritizes leads that provide the most informative anatomical views.

To integrate physiological and geometric information, we employ the Feature-wise Linear Modulation (FiLM) mechanism~\cite{perezFiLMVisualReasoning2018}, where the ECG-derived global feature $F_{ecg}$ is used as the condition to modulate the point cloud features $F_{pc}$ through affine transformations:
\begin{equation}
Z_{fused} = \gamma(F_{ecg}) \odot F_{pc} + \beta(F_{ecg})
\end{equation}
with \(\gamma(\cdot)\) and \(\beta(\cdot)\) as linear projections for scaling and shifting, and \(\odot\) denoting element-wise multiplication. This allows the network to selectively emphasize or suppress specific geometric channels.

Finally, the decoder reconstructs the spatial distribution of myocardial scars by decoupling these electro-anatomical features.
To train the network, we employ a composite loss function designed to optimize both segmentation accuracy and the plausibility of the anatomical structure:
\begin{equation}
\mathcal{L} = \mathcal{L}_{seg} + \lambda_{1}\mathcal{L}_{compact} + \lambda_{2}\mathcal{L}_{RV} + \lambda_{3}\mathcal{L}_{size}
\end{equation}
where $\mathcal{L}_{seg}$ is the segmentation cross-entropy loss, $\mathcal{L}_{compact}$ is the compactness loss~\cite{liEnablingCardiacDigital2024}, and $\mathcal{L}_{RV}$ and $\mathcal{L}_{size}$ are the penalty term for the right ventricle structure and the size constraint term, respectively. Since infarct regions are mostly located in the left ventricle, our pipeline considers only left ventricular infarcts for validation.  $\lambda$ represents the balancing weights for each loss term.

\section{Experiment}

\subsection{Experimental Setup}
\subsubsection{Dataset.}

We collected 230 multi-view cine MRI cases from the [Anonymous] hospital, of which 130 cases from 99 patients met the quality criteria for mesh reconstruction and electrophysiological simulation. Each case was generated with 17 simulated chronic MI scenarios, resulting in 2,210 simulated samples. The dataset was partitioned into training, validation, and test sets (70:15:15) using the inter-patient strategy to prevent subject-level data leakage. We also added real data in test set, containing 17 cases for external validation.

\subsubsection{Implementation Details.}
Our simulation uses a reaction-Eikonal solver with Mitchell-Schaeffer kinetics on 3D tetrahedral geometries with the edge resolution of 0.15\,cm. APD heterogeneity follows~\cite{campsHarnessing12leadECG2024} with $APD_{max/min} = 330.7/189.4$\,ms. Baseline conduction velocities (CVs) were set to 65, 51, 48\,cm/s along fiber, sheet and normal directions. Chronic MI is modeled by scaling CVs in scar and border zone to 10\% and 50\% of normal, respectively, and increasing APD by 30\% in the border zone to reflect chronic MI remodeling~\cite{zhouClinicalPhenotypesAcute2024}.
The inverse inference model was implemented in PyTorch and trained for 200 epochs using the Adam optimizer (batch size 4, weight decay $10^{-3}$) with an initial learning rate of $10^{-4}$, halved every 50 epochs via StepLR. Experiments were conducted on a machine with the Intel Xeon W7-2495X CPU and NVIDIA RTX A5500 GPU.

\subsection{Results}

\subsubsection{Sensitivity analysis of simulation.}
\begin{figure}[t]
    \centering
    \begin{subfigure}{0.48\linewidth}
        \centering
        \includegraphics[width=1.08\linewidth]{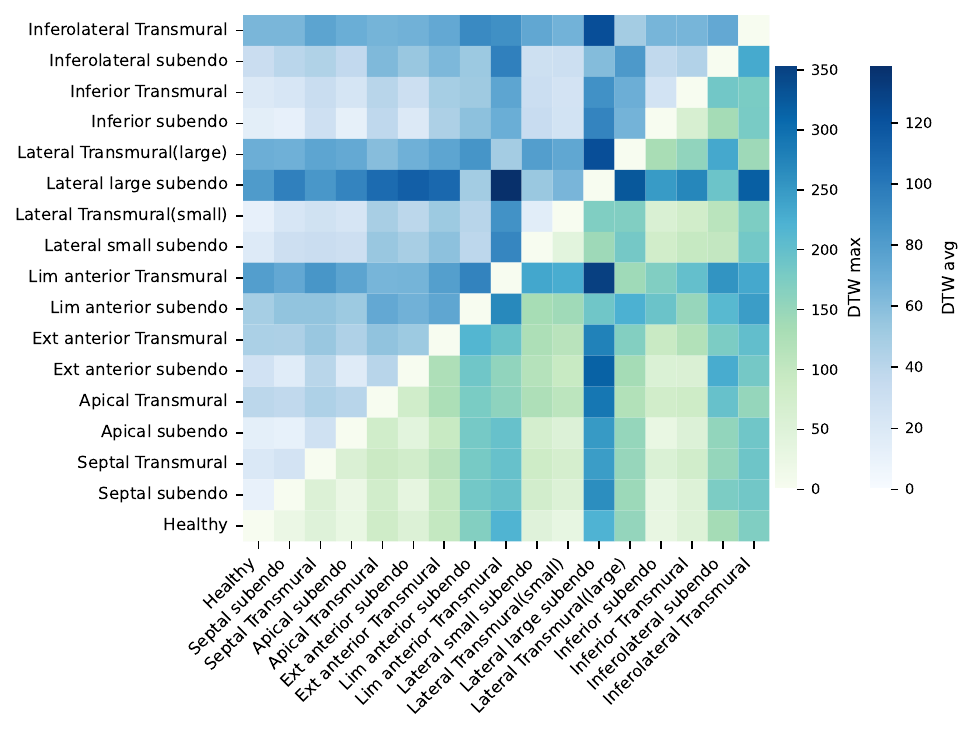}
        \caption{}
        \label{fig:SA-sub1}
    \end{subfigure}
    \hfill
    \begin{subfigure}{0.48\linewidth}
        \centering
        \includegraphics[width=0.86\linewidth]{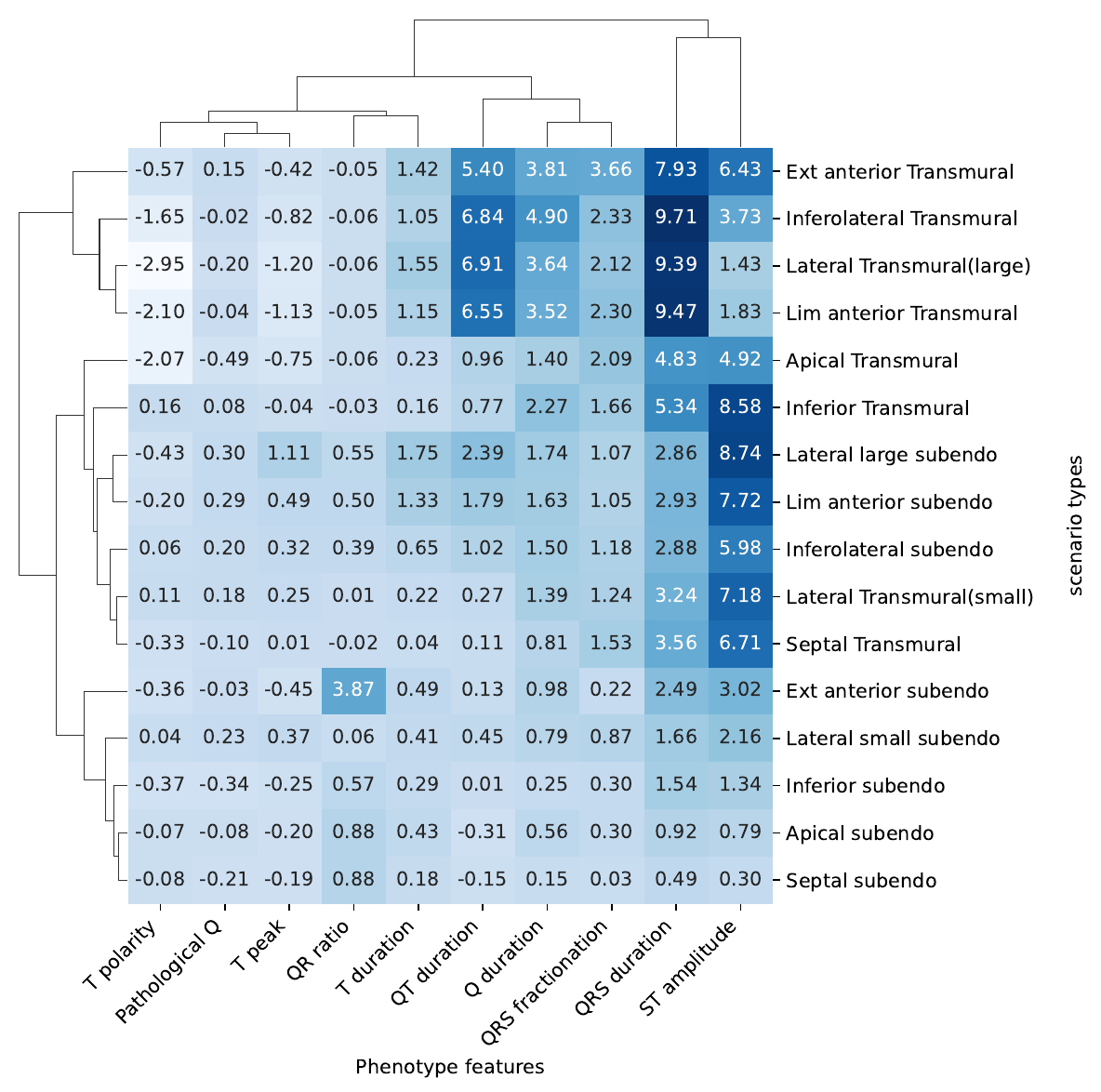}
        \caption{}
        \label{fig:SA-sub2}
    \end{subfigure}

    \caption{Sensitive analysis for synthetic MI scenarios. 
    (a) Mutual dissimilarity across each scenarios. ($DTW_{max}$ and $DTW_{avg}$ are maximum and average DTW across all leads).
    (b) Z-scores of ECG phenotype features for different MI scenarios compared to healthy. subendo: subendocardial.}
    \label{fig:SA}

\end{figure}

We conducted a sensitivity analysis to evaluate the distinguishability of the simulated ECGs.
In Fig.\ref{fig:SA-sub1}, we quantified global ECG dissimilarity using Dynamic Time Warping (DTW) distances between various scenarios.
While the average DTW distance exceeded 100 for distinct pairs, the subtle differences observed among other subtypes pose a challenge for the downstream inverse inference task, such as septal subendo.
Further analyzing specific features, Fig.\ref{fig:SA-sub2} details the deviation of specific ECG phenotypes between each MI scenarios and healthy baseline.
Hierarchical clustering reveals distinct patterns between transmural and subendocardial MI scenarios.
Notably, QRS duration and ST amplitude emerged as the most discriminative features, displaying significantly elevated Z-scores in the transmural cluster.

\subsubsection{Model Utility Analysis.}
Table \ref{tab:results_comparison} compares proposed PAA-Net with a VAE-based method~\cite{liEnablingCardiacDigital2024}. The proposed PAA-Net achieves the best overall performance, with mean Dice scores of 0.7391 for Scar and 0.5503 for BZ, and an AHA Score of 0.7375. Notably, our method shows substantial improvements in accurately localizing and quantifying small and irregular infarcts, such as Transmural Lateral(small). This demonstrates the robustness of our inverse inference framework across varying infarct sizes and locations. Furthermore, the external validation results demonstrate the superior generalizability of our model. The proposed PAA-Net consistently outperforms the VAE on unseen real-world data, achieving a higher Dice Scar score (0.2284 vs. 0.1540) and AHA Score (0.4490 vs. 0.3607). The AHA bullseye plot in Fig.~\ref{fig:external_visulization} shows that our network delineates the approximate MI locations even in a zero-shot real-world external dataset.

\subsubsection{Ablation Study.}
\begin{figure}[t]

  \centering
  \begin{minipage}[c]{0.4\textwidth}
    \centering
    \includegraphics[width=\textwidth]{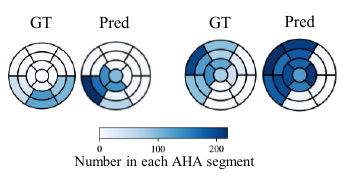}
    \caption{External validation visualization.}
    \label{fig:external_visulization}
  \end{minipage}
\hfill 
  \begin{minipage}[c]{0.55\textwidth}
    \centering
    \captionof{table}{Ablation studies.}
    \label{tab:ablation}
    \fontsize{8pt}{10pt}\selectfont 
    \begin{tabular}{l|cc|ccc}
      \toprule
      Methods  & A-SE & FiLM  & $\text{Dice}_{Scar}$ & $\text{Dice}_{BZ}$ & $\text{AHA}_{Score}$ \\
      \midrule
      Baseline & - & - & 0.7254           & 0.5315           & 0.7030           \\
      w/ CA    & - & -  & 0.7298           & 0.5305           & 0.7136           \\
        w/ A-SE  & $\checkmark$  & -   & {0.7309} & {0.5268} & {0.7280} \\
      w/ FiLM  & - & $\checkmark$   & {0.7373} & {0.5371} & {0.7375} \\

      \midrule
      PAA-Net  & $\checkmark$  & $\checkmark$   & \textbf{0.7391} & \textbf{0.5503} & \textbf{0.7375} \\
      \bottomrule
    \end{tabular}
  \end{minipage}

\end{figure}

Table \ref{tab:ablation} presents our ablation studies. Our baseline is a straightforward fusion model utilizing feature concatenation. The superior performance of the FiLM module suggests it provides an expressive mechanism for conditioning ECG representations on geometric features. This is critical for accurate scar reconstruction and surpasses the performance of cross-attention (w/ CA). Similarly, the A-SE module improves baseline performance, demonstrating its capability to selectively emphasize informative features from different leads. By combining these two modules, our PAA-Net achieves optimal performance across all three metrics, which are important for accurate MI localization.

\begin{table}[t]
  \centering
  \caption{Model performance in internal and external set.}
  \label{tab:results_comparison}

  \fontsize{8pt}{8pt}\selectfont
  \setlength{\tabcolsep}{3pt}

  \begin{tabular}{llcccccc}
\toprule
\multirow{2.5}{*}{Type} & \multirow{2.5}{*}{Location} & \multicolumn{2}{c}{$\text{Dice}_{Scar}$} & \multicolumn{2}{c}{$\text{Dice}_{BZ}$} & \multicolumn{2}{c}{$\text{AHA}_{Score}$} \\
\cmidrule(lr){3-4} \cmidrule(lr){5-6} \cmidrule(lr){7-8}
          &               & PAA-Net & VAE \cite{liEnablingCardiacDigital2024}    & PAA-Net & VAE \cite{liEnablingCardiacDigital2024}    &  PAA-Net & VAE \cite{liEnablingCardiacDigital2024} \\
\midrule
\multirow{8}{*}{\rotatebox{90}{Subendocardial}} 
          & Septal         & 0.4923 & 0.3777 & 0.4265 & 0.3474 & 0.5027 & 0.5524 \\
          & Apical         & 0.8417 & 0.8125 & 0.7701 & 0.7332 & 0.7635 & 0.7239 \\
          & Ext anterior   & 0.8406 & 0.8274 & 0.7542 & 0.7291 & 0.8515 & 0.8215 \\
          & Lim anterior   & 0.9004 & 0.9049 & 0.8068 & 0.7866 & 0.9501 & 0.9383 \\
          & Lateral (large)  & 0.8610 & 0.8483 & 0.7567 & 0.7502 & 0.7843 & 0.8261 \\
          & Lateral (small)  & 0.4779 & 0.4228 & 0.3264 & 0.2844 & 0.6047 & 0.5822 \\
          & Inferior       & 0.6265 & 0.5843 & 0.5276 & 0.5086 & 0.5998 & 0.6728 \\
          & Inferolateral  & 0.8399 & 0.8326 & 0.7463 & 0.7217 & 0.8680 & 0.8984 \\
\midrule
\multirow{8}{*}{\rotatebox{90}{Transmural}} 
          & Septal         & 0.6221 & 0.5900 & 0.3584 & 0.2695 & 0.6106 & 0.5321 \\
          & Apical         & 0.8778 & 0.8701 & 0.6024 & 0.5769 & 0.7737 & 0.7139 \\
          & Ext anterior   & 0.8546 & 0.8237 & 0.5679 & 0.5478 & 0.7922 & 0.7606 \\
          & Lim anterior   & 0.9057 & 0.9005 & 0.5975 & 0.5649 & 0.8608 & 0.8532 \\
          & Lateral (large)  & 0.8739 & 0.8711 & 0.5734 & 0.5264 & 0.8091 & 0.7209 \\
          & Lateral (small)  & 0.3151 & 0.1767 & 0.1413 & 0.1215 & 0.4767 & 0.2813 \\
          & Inferior       & 0.5777 & 0.5430 & 0.2749 & 0.2402 & 0.6632 & 0.6915 \\
          & Inferolateral  & 0.9190 & 0.9160 & 0.5739 & 0.5462 & 0.8893 & 0.8900 \\
\midrule
\multicolumn{2}{l}{\textbf{Mean}} & 
\textbf{0.7391} & 0.7064 & 
\textbf{0.5503} & 0.5159 & 
\textbf{0.7375} & 0.7162 \\
\multicolumn{2}{l}{\textbf{External validation}} & 
\textbf{0.2284} & 0.1540 & 
- & - & 
\textbf{0.4490} & 0.3607 \\
\bottomrule
\end{tabular}
\end{table}%

\subsubsection{Model Interpretability Analysis.}
\begin{figure}[t]
    \centering
    \includegraphics[width=0.95\textwidth]{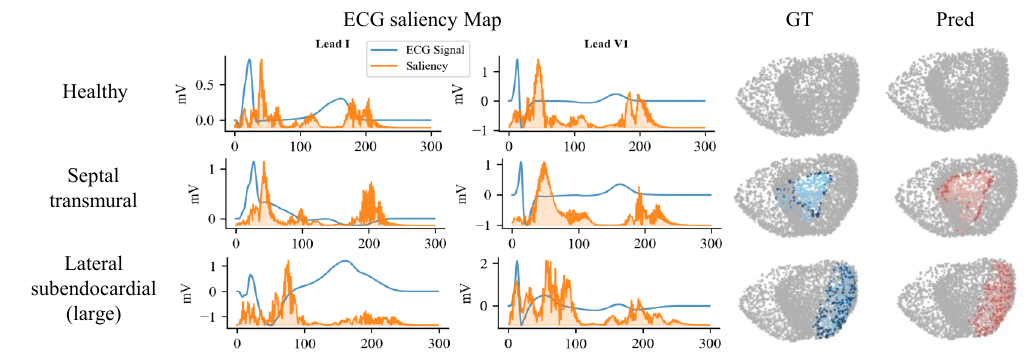}
    \caption{ECG saliency maps and qualitative infarct localization visualization.}
    \label{fig:visual}
\end{figure}

To validate that our PAA-Net framework captures clinically meaningful features, we visualize gradient-based saliency maps of the input ECG signals (Lead I and V1) in Fig.~\ref{fig:visual}. 
In infarction cases, the model exhibits strong attention on the QRS complex and ST-T segment, consistent with myocardial infarction affecting ventricular depolarization and repolarization. Notably, Cases of Septal transmural and Lateral subendocardial (large) show differences in their attention regions, reflecting the model’s sensitivity to variations in infarct location. In contrast, for the Healthy case, the model correctly predicts a clean mesh. Furthermore, the comparison between Lead I and V1 demonstrates the model's ability to adaptively weigh information from different spatial projections. The close overlap between the Ground Truth and Prediction on the 3D meshes further confirms that these learned ECG features are effectively translated into accurate spatial localization of the scar.

\section{Conclusion}
This paper introduces a comprehensive framework for MI localization using cine MRI and multi-lead ECG. The stochastic infarct synthesis strategy is developed to mimic real-world characteristics. The APD heterogeneity is incorporated to simulate a large virtual cohort of QRS-T waveforms. A novel PAA-Net framework is proposed to accurately localize MI of different locations, extents, and sizes. The model delineates the approximate location of the infarct even in external validation. Future work will focus on enhancing the realism of ECG simulations and integrating AI-based generative models. The infarct synthesis module will also be extended to simulate different time phases of MI progression.

\subsubsection{\ackname} This work was supported by Singapore National Medical Research Council Open Fund - Young Individual Research Grant (25-1321-A0001), and Singapore Ministry of Education Tier 1 grant (25-1097-P0001). M. Wang was partially supported by the China Scholarship Council (Grant No. 202506230034).
\subsubsection{\discintname}
The authors have no competing interests to declare that are relevant to the content of this article.

%
%
\bibliographystyle{splncs04}
\bibliography{references_used}

\end{document}